\documentclass{article}

\PassOptionsToPackage{numbers}{natbib}
\usepackage[preprint]{neurips_2019}

\usepackage[utf8]{inputenc} 
\usepackage[T1]{fontenc}    
\usepackage{hyperref}       
\usepackage{url}            
\usepackage{booktabs}       
\usepackage{amsfonts}       
\usepackage{microtype}      
\usepackage{graphicx}
\usepackage{amsmath}
\usepackage{color}
\usepackage{caption}
\usepackage{subfigure}
\usepackage[font=small,labelfont=bf]{caption}

\title{Remarks on Optimal Scores for Speaker Recognition}

\author{%
   Dong Wang \\
   CSLT@Tsinghua University \\
   \texttt{wangdong99@mails.tsinghua.edu.cn} \\
}

\begin{document}

\maketitle

\begin{abstract}

In this article, we first establish the theory of optimal scores for speaker recognition. Our analysis shows that
the minimum Bayes risk (MBR) decisions for both the speaker identification and speaker verification tasks can
be based on a normalized likelihood (NL). When the underlying generative model is a linear Gaussian,
the NL score is mathematically equivalent to the PLDA likelihood ratio,
and the empirical scores based on cosine distance and Euclidean distance can be seen as approximations of this linear Gaussian NL score under
some conditions. We discuss a number of properties of the NL score and perform a simple simulation experiment to demonstrate
the properties of the NL score.

\end{abstract}

\section{Introduction}
\label{sec:trend}

Speaker recognition research concerns two tasks: speaker identification (SI) that identify the true speaker
from a set of candidates, and speaker verification (SV) that tests if an alleged speaker is the true speaker. The performance of SI systems is evaluated by
identification rate (IDR), the percentage of the trials whose speakers are correctly identified. SV systems require a threshold to decide whether accepting the speaker or not
and the performance is evaluated by equal error rate (EER), to represent the trade-off between fail to accept and fail to reject~\cite{campbell1997speaker,reynolds2002overview,hansen2015speaker}.

Modern speaker recognition methods are based on the concept of
\emph{speaker embedding}, i.e., representing speakers by fixed-length continuous \emph{speaker vectors}. This embedding is traditionally based on statistical models, in particular the
i-vector model~\cite{dehak2011front}. Recently, deep learning methods gained much attention and embedding based on deep neural nets (DNN) becomes popular~\cite{ehsan14,li2017deep}.
A key component of the speaker embedding approach is how to score a trial.
Numerous empirical evidence has shown that the likelihood ratio (LR)
derived by probabilistic linear discriminant analysis (PLDA)~\cite{Ioffe06,prince2007probabilistic} works well in most situations, and when the computational resource is limited,
the cosine distance is a reasonable substitution.
In some circumstances in particular on SI tasks, the Euclidean distance can be used.

In this article, we revisit the scoring methods for speaker recognition from the perspective of
minimum Bayes risk (MBR). The analysis
shows that for both the SI and SV tasks, the MBR optimal score can be formulated as a single form $\frac{p_k(\pmb{x})}{p(\pmb{x})}$, which we
call a \emph{normalized likelihood (NL) score}. In the NL score,
$p_k(\pmb{x})$ is the likelihood term that represents the probability that the test utterance $\pmb{x}$ belongs to the target class $k$,
and $p(\pmb{x})$ is a normalization term that represents the probability that $\pmb{x}$ belongs to all possible classes.
We will show that the NL score is equivalent to PLDA LR, in the case where the speaker vectors are modeled by a linear Gaussian and the target class is
represented by finite enrollment utterances.
We will also show that under some conditions, the empirical scores based on cosine distance and Euclidean distance can be
derived from the linear Gaussian NL score.

It should be noted that the NL formulation is not new and may trace back to the LR scoring method with the Gaussian mixture model-Universal
background model (GMM-UBM) framework~\cite{Reynolds00}.
Within the speaker embedding framework, the NL form was derived by McCree et. al.~\cite{borgstrom2013discriminatively,mccree2017extended}
from the hypothesis test view (the one used for PLDA inference). Our derivation is based on the MBR decision theory, which
directly affirms the optimum of the NL score.

\section{MBR optimal decision and normalized likelihood}
\label{sec:mbr}

It is well known that an optimal decision for a classification task should minimize the Bayes risk (MBR):

\[
k^* = \arg\min_{k} \sum_j \ell_{jk} p(j|\pmb{x})
\]
\noindent where $\pmb{x}$ is the observation, $\ell_{jk}$ is the risk taken when classifying an observation from class $j$ to class $k$.
In the case where $\ell_{jk}$ is $0$ for $j=k$ and a constant $c$ for any $j \ne k$, the MBR decision is equal to selecting the class with the
largest posterior probability:

\[
k^* = \arg\max_{k} p(k|\pmb{x}).
\]

We call this result the \emph{MAP principle}.
We will employ this principle to derive the optimal score for the SI and SV tasks in speaker recognition.

\subsection{MBR optimal score for SI}

In the SI task, our goal is to test $K$ outcomes $\{H_k$: $\pmb{x}$ belongs to class $k\}$ and make the decision
which outcome is the most probable.
Following the MAP principle, the MBR optimal decision is to choose the $k$-th outcome that obtains the maximum posterior:

\[
k^* = \arg\max_k p(H_k|\pmb{x}) = \arg\max_k p_k(\pmb{x}) p(k),
\]
where $k$ indexes the classes, and $p_k(\pmb{x})$ represents the likelihood of $\pmb{x}$ in class $k$.
In most cases, there is no preference for
any particular class and so the prior $p(k)$ for each class $k$ shall be equal. We therefore have:

\[
k^* =  \arg\max_k p_k(\pmb{x}).
\]

\noindent It indicates that MBR optimal decisions can be conducted based on the likelihood $p_k(\pmb{x})$.
In other words, the likelihood is MBR optimal for the SI task.

\subsection{MBR optimal score for SV}

For the SV task, our goal is to test two outcomes and check which one is more probable: $\{$ $H_0$: $\pmb{x}$ belongs to class $k$; $H_1$: $\pmb{x}$ belongs
to any class other than $k$ $\}$.
Following the MAP principle, the MBR optimal decision should be based on the posterior $p(H_b|\pmb{x}): b=\{0,1\}$, if the risk for $H_0$ and $H_1$ is symmetric.
If the prior on each outcome is equal, we have:

\[
 p(H_b|\pmb{x}) = \frac{p(\pmb{x}|H_b)}{p(\pmb{x}|H_0) + p(\pmb{x}|H_1)}.
\]

\noindent Since $p(H_0|\pmb{x}) + p(H_1|\pmb{x})=1$, the decision can be simply made according to  $p(H_0|\pmb{x})$:

\[
b^* = \begin{cases}
 0 & if \ p(H_0|\pmb{x}) \ge 0.5\\
 1 & if \ p(H_0|\pmb{x}) < 0.5.
\end{cases}
\]

\noindent In practice, by setting an appropriate threshold on $p(H_0|\pmb{x})$, one can deal with different priors and
risk on $H_0$ and $H_1$. We highlight that for any class $k$, this threshold is only related to the prior and risk.
This is important as it means that based on $p(H_0|\pmb{x})$, MBR optimal decisions
can be made \emph{simultaneously} for all the classes by setting a global threshold.
A simple case is to set the threshold to 0.5 when the risk is symmetric
and the priors are equal. In summary, $p(H_0|\pmb{x})$ is MBR optimal for the SV task.

Note that when computing the posterior $p(H_0|\pmb{x})$, $p(\pmb{x}|H_0)$ is exactly the likelihood $p_k(\pmb{x})$, and $p(\pmb{x}|H_1)$ summarizes the
likelihood of all possible classes except the class $k$.
In most cases, an SV system is required to deal with any unknown
class, and so the class space is usually assumed to be continuous. To simplify the presentation, we will assume each class
being uniquely represented by the mean vector $\pmb{\mu}$ and $p(\pmb{\mu})$ is continuous. In this case,
the contribution of each class is infinitely small and so $p(\pmb{x}|H_1)$ is exactly the marginal
distribution (or evidence) $p(\pmb{x})=\int p(\pmb{x}|\pmb{\mu}) \rm{d} \pmb{\mu}$.\footnote{One may argue
that $p(\pmb{x})$ involves the quantity $p_k(\pmb{x})$, and so is not accurately $p(\pmb{x}|H_1)$. This is not true
however, as the contribution of $p_k(\pmb{x})$ to $p(\pmb{x})$ is zero if $p(\pmb{\mu})$ is continuous. This also means that
the likelihood that $\pmb{x}$ belongs to all classes equals to the likelihood that $\pmb{x}$ belongs to all classes other than $k$.
} We therefore obtain the
MBR optimal score for SV:

\[
p(H_0|\pmb{x}) = \frac{p_k(\pmb{x})}{p_k(\pmb{x}) + p(\pmb{x})}.
\]

\subsection{Normalized likelihood}

Note that for the SV task, the posterior $p(H_0|\pmb{x})$ is determined by the ratio $p(\pmb{x}|H_0)/p(\pmb{x}|H_1)$, which is essentially
the class-dependent likelihood $p_k(\pmb{x})$ normalized by the class-independent likelihood $p(\pmb{x})$.
We therefore define the normalized likelihood (NL) as:

\begin{equation}
\label{eq:nl}
NL (\pmb{x}|k)= \frac{p(\pmb{x}|H_0)}{p(\pmb{x}|H_1)} = \frac{p_k(\pmb{x})}{p(\pmb{x})}.
\end{equation}

\noindent Note that the NL is linked to the posterior $p(H_0|\pmb{x})$ by a monotone function:

\[
NL (\pmb{x} |k ) = \frac{p(H_0|\pmb{x})}{1 - p(H_0|\pmb{x})}.
\]

\noindent Since the posterior $p(H_0|\pmb{x})$ is MBR optimal for the SV task, the NL is also MBR optimal as a threshold on
$p(H_0|\pmb{x})$ that leads to (global) MBR decisions can be simply transformed to a threshold on the NL, by which
the same MBR decisions can be achieved.

Interestingly, the NL score is also MBR optimal for the SI task. This is because the normalization term $p(\pmb{x})$ is the same for all classes
in the SI task,
so the decisions made based on the NL score is equal to those based on the likelihood $p_k(\pmb{x})$. Since the likelihood is MBR optimal for
the SI task, the NL score is MBR optimal for the SI task as well. \textbf{We therefore conclude that the NL score is MBR optimal for both
the SI and the SV tasks.} It should be noted that the NL form Eq. (\ref{eq:nl}) is a high-level definition and it can be implemented in a flexible way.
In particular, $p_k(\pmb{x})$ and $p(\pmb{x})$
can be any models that produce the class-dependent and class-independent likelihoods respectively.

Finally, NL is not new for speaker recognition. It is essentially the likelihood ratio (LR) that has been employed
for many years since the GMM-UBM regime, where the score is computed by $\frac{p_{GMM}(\pmb{x})}{p_{UBM}(\pmb{x})}$.
We use the term NL instead of LR in this paper in order to: (1) highlight the different roles of the numerator $p_k(\pmb{x})$ and
the denominator $p(\pmb{x})$ in the ratio; (2)
discriminate the normalization-style LR (used by NL) and the comparison-style LR, e.g., the one used by PLDA inference that compares the
likelihoods that a group of samples are generated from the same and different classes.

\section{NL score with linear Gaussian model}
\label{sec:gauss}

Although the NL framework allows flexible models for the class-dependent and class-independent likelihoods, linear
Gaussian model is the most attractive due to its simplicity. We derive the NL score with this model, for the case
(1) the class means have been known and (2) the class means are unknown and have to be estimated from enrollment data.

\subsection{Linear Gaussian model}

We shall assume a simple linear Gaussian model for the speaker vectors that we will score:

\begin{equation}
\label{eq:pmu}
p(\pmb{\mu}) = N(\pmb{\mu}; \pmb{0},  \mathbf{I} \pmb{\epsilon}^2 )
\end{equation}

\begin{equation}
\label{eq:px-mu}
p(\pmb{x}|\pmb{\mu}) = N(\pmb{x}; \pmb{\mu}, \sigma^2 \mathbf{I}),
\end{equation}

\noindent where $\pmb{\mu} \in R^D$ represents the means of classes and $\pmb{x} \in R^D$ represents observations, and $\pmb{\epsilon}^2 \in (R^+)^D$ and $\sigma^2 \in R^+$
represent the between-class and within-class
variances respectively. Applied to speaker recognition, $\pmb{\epsilon}$ and $\sigma$ represent the
between-speaker and within-speaker variances respectively. We highlight that any linear Gaussian model can be transformed into this simple form (i.e., isotropic within-class
covariance and diagonal between-class covariance) by a linear transform such as full-dimensional linear discriminant analysis (LDA),
and this linear transform will not change the identification and verification results as we will show in
Section~\ref{sec:remark}. Therefore, study with the simple form Eq. (\ref{eq:pmu}) and Eq. (\ref{eq:px-mu}) is
sufficient for us to understand the behavior of a general linear Gaussian model with complex covariance matrices.

With this model, it is easy to derive the marginal probability $p(\pmb{x})$ and the posterior probability $p(\pmb{\mu}|\pmb{x})$ as follows:

\begin{equation}
\label{eq:px}
p(\pmb{x}) = N(\pmb{x}; \pmb{0}, \mathbf{I} (\pmb{\epsilon}^2 + \sigma^2) )
\end{equation}

\begin{equation}
\label{eq:pmu-x}
 p(\pmb{\mu}|\pmb{x}) = N(\pmb{\mu}; \frac{\pmb{\epsilon}^2}{\pmb{\epsilon}^2 + \sigma^2} \pmb{x}, \mathbf{I} \frac{\sigma^2 \pmb{\epsilon}^2}{\pmb{\epsilon}^2 + \sigma^2} ),
\end{equation}
\noindent where all the operations between vectors are element-wised and appropriate dimension expansion has been assumed, e.g., $\pmb{\epsilon}^2 + \sigma^2 = \pmb{\epsilon}^2 + [\sigma^2 \ ... \ \sigma^2 ]^T$.

If the observations are more than one, the posterior probability has the form:

\begin{equation}
\label{eq:pmu-xx}
 p(\pmb{\mu}|\pmb{x}_1,...,\pmb{x}_n) = N(\pmb{\mu}; \frac{n \pmb{\epsilon}^2}{n\pmb{\epsilon}^2 + \sigma^2} \bar{\pmb{x}},  \mathbf{I} \frac{ \sigma^2 \pmb{\epsilon}^2}{n\pmb{\epsilon}^2 + \sigma^2} ),
\end{equation}

\noindent where $\bar{\pmb{x}}$ is the average of the observations. These equations will be extensively used in the following sections.

\subsection{Case 1: class means are known}

In this case, we assume that the class means are known. This is equivalent to say that each class is represented by infinite enrollment data.

\textbf{NL/Euclidean/Cosine scores for SI}

For the SI tasks, decisions based on  the NL score and the likelihood $p_k(\pmb{x})$ are the same and both are MBR optimal. With the linear Gaussian model, the likelihood is:

\begin{equation}
\label{eq:pk}
p_k(\pmb{x}) = N(\pmb{x}; \pmb{\mu}_k, \sigma^2 \mathbf{I}).
\end{equation}

A simple rearrangement shows that:

\[
\log p_k(\pmb{x}) =  -\frac{1}{2 \sigma^2} ||\pmb{x} - \pmb{\mu}_k||^2  + const
\]

Since the variance $\sigma$ is the same for all classes, the MBR decision can be equally based on the Euclidean distance, e.g.,

\[
s_{e} = ||\pmb{x} - \pmb{\mu}_k||^2,
\]

\noindent where we use $s_{e}$ to denote the score based on the Euclidean distance. In short, the Euclidean score is MBR optimal
for the SI task when the class means are known.

Next, we will show that in a high-dimensional space, the Euclidean distance is well approximated by
the cosine distance, under the linear Gaussian assumption.

\begin{figure}[htbp]
\centering\includegraphics[width=0.5\linewidth]{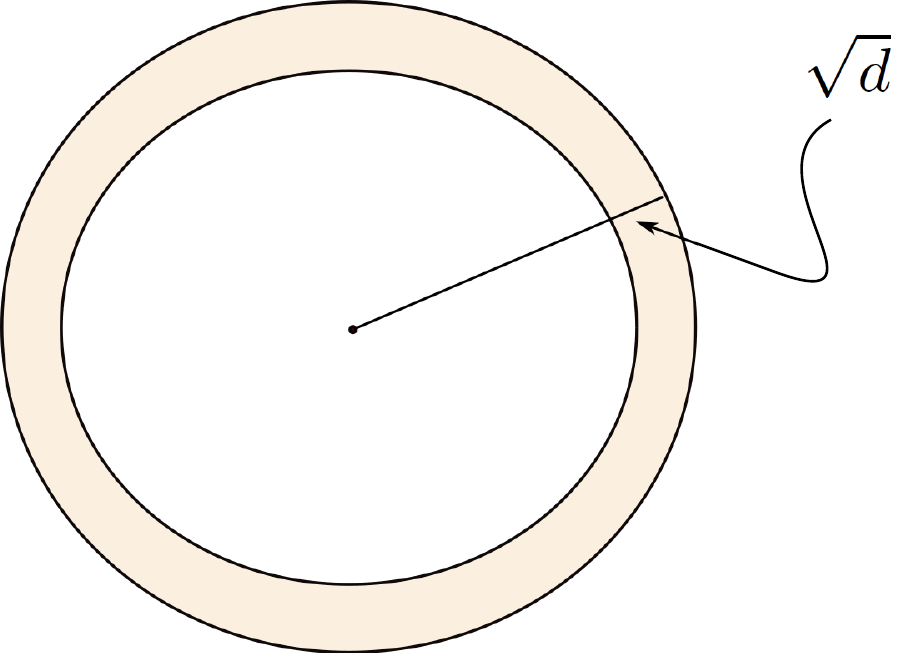}
\caption{Gaussian annulus theorem~\cite{blum2020foundations}: for a $d$-dimensional multi-variant Gaussian with unit variance in all directions,
for any $\beta \le \sqrt{d}$, all but at most $3e^{-c\beta^2}$
of the probability mass lies within the annulus $\sqrt{d}-\beta \le ||\emph{x}|| \le \sqrt{d} + \beta$, where $c$ is a fixed positive constant.
The color region shown in the figure represents the annulus. Rigid proof can be found in~\cite{blum2020foundations}.
}
\label{fig:shell}
\end{figure}

First notice that the Gaussian annulus theorem~\cite{blum2020foundations} states that for a $d$-dimensional Gaussian distribution
with the same variance $\epsilon$ in each direction, nearly all the probability mass is concentrated in a thin annulus of width $O(1)$ at radius $\sqrt{d}\epsilon $,
as shown in Figure~\ref{fig:shell}.
This slightly anti-intuitive result indicates that in a high-dimensional space, most of the samples from a Gaussian tend to be in the same length.
Rigid proof for this theorem can be found in~\cite{blum2020foundations}.

Now we rewrite the Euclidean score as follows:

\[
s_{e} = ||\pmb{x}||^2 + ||\pmb{\mu}_k||^2 - 2 \cos (\pmb{x}, \pmb{\mu}_k) ||\pmb{x}|| \ ||\pmb{\mu}_k||,
\]

\noindent since $||\pmb{\mu}_k|| \approx \sqrt{d}\epsilon$, $\cos (\pmb{x}, \pmb{\mu}_k)$ will be the
only term that discriminates the probability that $\pmb{x}$ belongs to different class $k$. This leads to the cosine score:

\[
s_{c} = \cos (\pmb{x}, \pmb{\mu}_k).
\]

This result provides the theoretical support for the cosine score. It should be noted that this approximation is only valid for high-dimensional data, and
the class means must be from a Gaussian with a zero mean. Therefore, data centralization is important for cosine scoring.

\textbf{NL/Euclidean/Cosine scores for SV}

For the SV task, the MBR optimal decision should be based on the NL score. With the linear Gaussian model, one can easily show that:

\begin{equation}
\label{eq:nl-known}
\log NL (\pmb{x}|k) = -\frac{1}{2 \sigma^2} ||\pmb{x} - \pmb{\mu}_k||^2 + \frac{1}{2}||\frac{\pmb{x}}{\sqrt{\pmb{\epsilon}^2 + \sigma^2}} ||^2 + \ const.
\end{equation}

A simple rearrangement shows that:

\begin{eqnarray}
\log NL (\pmb{x} |k ) &=& -\frac{1}{2\sigma^2} (||\pmb{x}||^2 + ||\pmb{\mu}_k||^2 - 2 \cos (\pmb{x}, \pmb{\mu}_k) ||\pmb{x}|| \ ||\pmb{\mu}_k||) +  \frac{1}{2}||\frac{\pmb{x}}{\sqrt{\pmb{\epsilon}^2 + \sigma^2}} ||^2 + \ const \nonumber \\
        &\propto& - \Big\{ ||\frac{\pmb{\epsilon}}{\sqrt{\sigma^2+\pmb{\epsilon}^2}} \pmb{x}||^2  + ||\pmb{\mu}_k||^2 - 2 \cos (\pmb{x}, \pmb{\mu}_k) ||\pmb{x}|| \ ||\pmb{\mu}_k||  \Big\} . \label{eq:nl-known-cosine}
\end{eqnarray}

It can be seen that if the within-class variance $\sigma^2$ is significantly larger than the between-class variance $\pmb{\epsilon}^2$ (we refer to element-based comparison here and after),
the $\log NL$ will significantly depart from the Euclidean distance, but more closely related to
the cosine distance.
Essentially, if we admit that both $||\pmb{x}||^2$ and $||\pmb{\mu}_k||^2$ tend to be constant due to the Gaussian annulus theorem, the cosine score will be a
good approximation for the optimal $\log NL$. Conversely, if the between-class variance $\pmb{\epsilon}^2$ is sufficient larger than the within-class
variance $\sigma^2$, it can be well approximated by the Euclidean score.

\subsection{Case 2: class means are unknown}

In the pervious section, we have supposed that the class means are known precisely. In real scenarios, however, this is not possible. We usually have only a few
enrollment samples (e.g., less than 3) to represent the class, and the SI or SV evaluation should be based on these representative samples. In this case, the class means are
unknown and have to be estimated from the enrollment data, leading to uncertainty that must be taken into account during scoring.

\textbf{NL/Euclidean/Cosine scores for SI}

Firstly consider the MBR optimal decision for SI. As in the known-mean scenario, we compute the likelihood for the $k$-th class:

\[
p_k(\pmb{x}) = N(\pmb{x}; \pmb{\mu}_k, \sigma^2 \mathbf{I}).
\]

\noindent An important difference here is  that $\pmb{\mu}_k$ is unknown, and so has to be estimated from the enrollment samples belonging to the same class. Denoting
these samples by $\pmb{x}^k_1,...\pmb{x}^k_{n_k}$ and their average by $\bar{\pmb{x}}_k$,  we have the posterior probability for the class mean $\pmb{\mu}_k$,
according to Eq.~(\ref{eq:pmu-xx}):

\begin{equation}
\label{eq:posterior-unknown}
p(\pmb{\mu}_k |\pmb{x}^k_1,...\pmb{x}^k_{n_k} ) = N(\pmb{\mu}_k ;  \frac{n_k\pmb{\epsilon}^2}{n_k\pmb{\epsilon}^2 + \sigma^2} \bar{\pmb{x}}_k, \mathbf{I} \frac{\sigma\pmb{\epsilon}^2}{n_k\pmb{\epsilon}^2 + \sigma^2}  ).
\end{equation}

The likelihood $p_k(\pmb{x})$ can therefore be computed by marginalizing over $\pmb{\mu}_k$, according to this posterior. Following Eq.(\ref{eq:px}), we have:

\begin{eqnarray}
p_k(\pmb{x}) &=&p(\pmb{x}|\pmb{x}^k_1,..., \pmb{x}^k_{n_k}) \nonumber \\
             &=& \int p(\pmb{x}| \pmb{\mu}_k) p(\pmb{\mu}_k|\pmb{x}^k_1,...\pmb{x}^k_{n_k}) \rm{d} \pmb{\mu}_k \nonumber \\
             &=& N(\pmb{x}; \frac{n_k\pmb{\epsilon}^2}{n_k\pmb{\epsilon}^2 + \sigma^2} \bar{\pmb{x}}_k, (\sigma^2 + \mathbf{I} \frac{\sigma\pmb{\epsilon}^2}{n_k\pmb{\epsilon}^2 + \sigma^2}) ). \nonumber
\end{eqnarray}

Note that with the class mean uncertainty, the Euclidean score is not MBR optimal anymore.
If the number of enrollment observations are the same for all classes,
the likelihood is exclusively determined by the class mean $\pmb{\mu}_k$. In this case, an amended version of the Euclidean score is optimal,
where the class mean is computed by  $\frac{n_k\pmb{\epsilon}^2}{n_k\pmb{\epsilon}^2 + \sigma^2}\pmb{\mu}_k$.
Note that the scale $\frac{n_k\pmb{\epsilon}^2}{n_k\pmb{\epsilon}^2 + \sigma^2}$ has been applied to compensate for the uncertainty of the
maximum-likelihood mean estimation $\pmb{\mu}_k$.
Intuitively, a smaller $n$ or a larger $\sigma^2/\pmb{\epsilon}^2$ lead to more uncertainty, so the compensation term will be more significant.
With more enrollment samples, the compensation term will converge to one, and the standard Euclidean score is recovered.

Another observation is that the scale compensation on $\pmb{\mu}_k$ does not change its direction. This implies that the cosine score does not
need any amendment to account for the uncertainty. However, it does not mean that the cosine score is not impacted by the class mean uncertainty;
it just means that the cosine score is not impacted as much as the Euclidean score.

\textbf{NL/Euclidean/Cosine scores for SV}

Now we normalize the score $p_k(\pmb{x})$ to make it suitable for SV, by introducing
a normalization term $p(\pmb{x})$:

\begin{eqnarray}
\label{eq:nl-unknown}
NL (\pmb{x} |k)= \frac{p(\pmb{x}|\pmb{x}_1, ..., \pmb{x}_{n_k})} {p(\pmb{x})}.
\end{eqnarray}

Note that the normalization term $p(\pmb{x})$ is not impacted by the mean uncertainty, and
therefore remains the same value as in the known-mean scenario. A simple computation shows that:

\begin{equation}
\label{eq:nl-unknown-log}
\log NL (\pmb{x} |k ) \propto -||\frac{\pmb{x} -  \tilde{\pmb{\mu}}_k}{\sqrt{\sigma^2 + \frac{\pmb{\epsilon}^2 \sigma^2}{n_k \pmb{\epsilon}^2 + \sigma^2}}} ||^2 + ||\frac{\pmb{x}}{\sqrt{\pmb{\epsilon}^2 + \sigma^2}} ||^2,
\end{equation}
\noindent where we have defined:

\[
\tilde{\pmb{\mu}}_k = \frac{n_k\pmb{\epsilon}^2}{n_k\pmb{\epsilon}^2 + \sigma^2} \bar{\pmb{x}}_k.
\]

To compare with the Euclidean score and the cosine score, Eq.~(\ref{eq:nl-unknown}) can be reformulated to:

\begin{equation}
\label{eq:nl-unknown-cosine}
\log NL (\pmb{x} |k ) \propto - \Big\{\frac{n_k \pmb{\epsilon}^4}{(\sigma^2 +\pmb{\epsilon}^2)(n_k \pmb{\epsilon}^2 + \sigma^2)}||\pmb{x}||^2  + ||\tilde{\pmb{\mu}}_k||^2 - 2 \cos (\pmb{x}, \tilde{\pmb{\mu}}_k) ||\pmb{x}|| \ ||\tilde{\pmb{\mu}}_k||  \Big\}.
\end{equation}

\noindent It can be seen that if the between-class variance $\pmb{\epsilon}$ is significantly smaller than the within-class variance $\sigma$,
the first two terms on the right hand side of Eq.~(\ref{eq:nl-unknown-cosine}) tend to be small and $\log NL$ can be approximated by the cosine score.
On the opposite, if the between-class variance $\pmb{\epsilon}$ is significantly larger than the within-class variance $\sigma$,
the amended Euclidean score will be a good approximation. Finally, if $n_k$ is sufficiently large, Eq.~(\ref{eq:nl-unknown-cosine})
will fall back to Eq.~(\ref{eq:nl-known-cosine}) of the know-mean case.

\section{Remarks on properties of NL score}
\label{sec:remark}

\textbf{Remark 1: Equivalent to PLDA LR}

The NL score based on the linear Gaussian model and unknown class means is equivalent to the PLDA LR~\cite{Ioffe06,prince2007probabilistic}.
PLDA assumes the same linear Gaussian model, but uses the following likelihood ratio
as the score:

\[
LR_{PLDA} (\pmb{x}) = \frac{p(\pmb{x},\pmb{x}_1,...,\pmb{x}_n \ from \ the \ same \ class ) }{p (\pmb{x} \ from \ a \ unique \ class) p (\pmb{x_1, ..., x_n} \ from \ a \ unique \ class)}
\]

Note that this likelihood ratio is different from the likelihood ratio of the NL score in Eq. (\ref{eq:nl}). The PLDA LR can be formally represented by:

\[
LR_{PLDA} (\pmb{x}) = \frac{p(\pmb{x},\pmb{x}_1,...,\pmb{x}_n)}{p (\pmb{x}) p (\pmb{x}_1, ..., \pmb{x}_n)}
\]

\noindent where $p(\pmb{x}_1, ..., \pmb{x}_n)$ denotes the probability that $\pmb{x}_1, ..., \pmb{x}_n$ belong to \emph{the same but an unknown} class. In principle,
this quantity can be computed by marginalizing over the class mean:

\[
p(\pmb{x}_1, ..., \pmb{x}_n) = \int p(\pmb{x}_1, .., \pmb{x}_n | \pmb{\mu}) p(\pmb{\mu}) \rm{d} \pmb{\mu}.
\]

A simple re-arrangement shows that:

\[
LR_{PLDA} (\pmb{x}) = \frac{p(\pmb{x}| \pmb{x}_1, ..., \pmb{x}_n)}{p(\pmb{x})} = \frac{\int p(\pmb{x}|\pmb{\mu}) p(\pmb{\mu}| \pmb{x}_1, ..., \pmb{x}_n) \rm{d} \pmb{\mu}}{p(\pmb{x})},
\]

\noindent where we have divided the numerator $p(\pmb{x},\pmb{x}_1,...,\pmb{x}_n)$ by $p(\pmb{x}_1, ..., \pmb{x}_n)$, which converts the marginal distribution $p(\pmb{x}, \pmb{x}_1, ..., \pmb{x}_n)$
to the conditional distribution $p(\pmb{x}| \pmb{x}_1, ..., \pmb{x}_n)$. By this change, the numerator is the likelihood of $\pmb{x}$ belonging to the
class represented by $\pmb{x}_1, ..., \pmb{x}_n$, and the denominator is the likelihood $\pmb{x}$ belonging to any class. This is exactly the normalized
likelihood in Eq.(\ref{eq:nl-unknown}). We therefore conclude that the PLDA LR is an NL where the underlying probabilistic model is linear Gaussian and the
class means are estimated from finite enrollment data.
\textbf{Since the NL score is MBR optimal for both SI and SV tasks, an immediate conclusion is that the PLDA LR is also MBR optimal for the two tasks.}
Note that the NL form of the PLDA LR was discussed by McCree et. al. ~\cite{borgstrom2013discriminatively,mccree2017extended}.

Compared to PLDA LR, NL possesses some attractive properties and brings some interesting merits.
A particular merit is that NL decouples the score computation into three
steps: posterior computation based on enrollment data, likelihood computation for the test data based on the posterior, and
normalization based on a global model. This offers an interesting correspondence between the scoring model and the scoring process.
We therefore can investigate the behavior of each component and design fine-grained treatment for real-life imperfection.

\textbf{Remark 2: Invariance with invertible transform}

Suppose an invertible transform $g$ on $\pmb{x}$, and
the probabilities on $\pmb{x}$ and $g(\pmb{x})$ are $p$ and $p'$ respectively.
According to the principle of distribution transformation for continuous variables~\cite{rudin2006real},
$p$ and $p'$ has the following relation:

\begin{equation}
\label{eq:nf}
p'(g(\pmb{x})) = p(\pmb{x}) |\det \frac{\partial{g}^{-1}(\pmb{x})}{\partial{\pmb{x}}}|,
\end{equation}
\noindent where the second term is the absolute value of the determinant of the Jacobian matrix of $g^{-1}$, the inverse transform of $g$.
This term reflects the change of the volume with the transform, and is often called the entropy term and denoted by $J(\pmb{x})$.

For the marginal distribution $p(\pmb{x}_1, ..., \pmb{x}_n)$ where $\pmb{x}_1, ..., \pmb{x}_n$ are drawn from the same but an unknown
class, one can compute the distribution by:

\begin{eqnarray}
p(\pmb{x}_1, ..., \pmb{x}_n) &=& \int \prod_{i=1}^{n} p(\pmb{x}_i | \pmb{\mu}) p(\pmb{\mu}) \rm{d} \pmb{\mu} \nonumber \\
                             &\approx& \sum_{j} \prod_{i=1}^{n} p(\pmb{x}_i | \pmb{\mu}_j) p(\pmb{\mu}_j) \Delta(\pmb{\mu}_j) \nonumber
\end{eqnarray}
\noindent  where we have divide the $\pmb{\mu}$ space into a large amount of small areas $\{\Delta(\pmb{\mu}_j)\}$ with centers $\{\pmb{\mu}_j\}$.
The approximation will approach to be accurate when the number of small areas is infinite.
With the transform $g$, we have:

\[
p'(g(\pmb{x}_1), ..., g(\pmb{x}_n)) \approx \sum_{j} \prod_{i=1}^{n} p'(g(\pmb{x}_i) | \pmb{\mu}^g_j) p'(g(\pmb{\mu}_j)) \Delta(g(\pmb{\mu}_j)) \nonumber
\]
\noindent where $\pmb{\mu}^g_j$ represents the mean of the class centered at $\pmb{\mu}_j$ after the transform.
Moreover, the transform $g$ does not change the probability within $\Delta(\pmb{\mu}_j)$, which means:
\[
p(\pmb{\mu}_j) \Delta(\pmb{\mu}_j) = p'(g(\pmb{\mu}_j)) \Delta(g(\pmb{\mu}_j)).
\]

Putting all the pieces together, we have:

\begin{eqnarray}
p'(g(\pmb{x}_1), ..., g(\pmb{x}_n)) &\approx& \sum_{j} \prod_{i=1}^{n} p'(g(\pmb{x}_i) | \pmb{\mu}^g_j)) p(\pmb{\mu}_j) \Delta(\pmb{\mu}_i) \nonumber \\
                                   &=&  \sum_{j} \prod_{i=1}^{n} J (\pmb{x}_i) \prod_{i=1}^{n} p(\pmb{x}_i | \pmb{\mu}_j) p(\pmb{\mu}_j)  \Delta(\pmb{\mu}_i), \nonumber
\end{eqnarray}
\noindent where we have applied the rule of the distribution transform shown in Eq.~(\ref{eq:nf}). Let the size of $\{\Delta(\pmb{\mu}_j)\}$ to be infinite,
we have the marginal distribution in the space induced by transform $g$:

\begin{equation}
p'(g(\pmb{x}_1), ..., g(\pmb{x}_n)) = \prod_{i=1}^n J(\pmb{x}_i) p(\pmb{x}_1, ..., \pmb{x}_n). \nonumber
\end{equation}

Substituting back to the NL score, we obtain the invariance of the NL score under an invertible transform:

\begin{eqnarray}
NL(g(\pmb{x})|g(\pmb{x}_1), ..., g(\pmb{x}_{n_k})) &=& \frac{p'(g(\pmb{x}),g(\pmb{x}_1), ..., g(\pmb{x}_{n_k})}{p'(g(\pmb{x})) p'(g(\pmb{x}_1), ..., g(\pmb{x}_{n_k})) } \nonumber \\
                                                   &=& \frac{ J(\pmb{x}) \prod_{i=1}^n J(\pmb{x}_i)  p(\pmb{x},\pmb{x}_1, ..., \pmb{x}_{n_k})}{ \big\{ J(\pmb{x}) p(\pmb{x}) \big\} \big\{ \prod_{i=1}^n J(\pmb{x}_i) p(\pmb{x}_1, ..., \pmb{x}_{n_k})\big\} } \nonumber \\
                                                   &=& NL(\pmb{x}|\pmb{x}_1, ..., \pmb{x}_{n_k}) \nonumber
\end{eqnarray}
\noindent where we have employed the PLDA LR form to represent the NL score.

The above derivation indicates that the NL score can be computed in a transformed space induced by an invertible
transform. Among all the possible invertible transforms, the full-dimension LDA is particularly attractive. It can simultaneously
diagonalize the within-class and between-class covariances and regulate the
within-class covariance to be identity. \textbf{We therefore do not need consider the general form of
distributions when investigating the properties of the NL score,
instead just focusing on the simple form with diagonal covariances, as we did in the previous sections}.

\textbf{Remark 3: Dimensionality is important}

Let's investigate the benefit of a high-dimensional space. It has been shown~\cite{blum2020foundations} that the
distance of two random samples from a $n$-dimensional Gaussian with variance $\epsilon^2$ in all directions has a large
probability to be:

\[
||\pmb{x} - \pmb{y}|| = \sqrt{2d}\epsilon \pm O(1).
\]

Consider the class means are random samples of a Gaussian with variance $\epsilon^2$, and each class is a Gaussian with variance $\sigma^2$.
Due to the Gaussian annulus theorem,
the samples of each class will concentrate in the annulus of radius $\sqrt{d}\sigma$. Since the distance of two class means
has a large probability to be $\sqrt{2 d}\epsilon$, it is easy to conclude that if $2\sigma < \epsilon$, there will be a large probability that
most of the classes are well separated.

More careful analysis shows a better bound. Considering two samples from two different classes respectively, it
shows that their distance tend to be $\sqrt{\Delta^2 + 2\sigma^2 d \pm O(\sqrt{d}\sigma)}$, where $\Delta$ is the distance
of these two classes, and $\sigma^2$ is the variance of each class~\cite{blum2020foundations}. Since the samples from the
same class tends to be $\sqrt{2 d}\sigma$, one can show if $\Delta^2 \ge O(\sqrt{d}\sigma)$, there will be a large probability
to identify if two samples are from the same class or different classes. If the class means are sampled from a Gaussian with variance $\epsilon$,
we will have $\Delta^2 \approx 2 \epsilon^2 d$. One can easily derive that if $\sigma^2 \le O(\epsilon^4 d)$, sample pairs from two classes
can be well differentiated from sample pairs from the same class. Note the condition depends on $d$, which means that
with a higher dimension, classes with larger variances can be separated with a large probability.
In other words, classes in higher dimensional space tend to be more separable.

\textbf{Remark 4: Direction is important}

\begin{figure}[htbp]
\centering\includegraphics[width=0.5\linewidth]{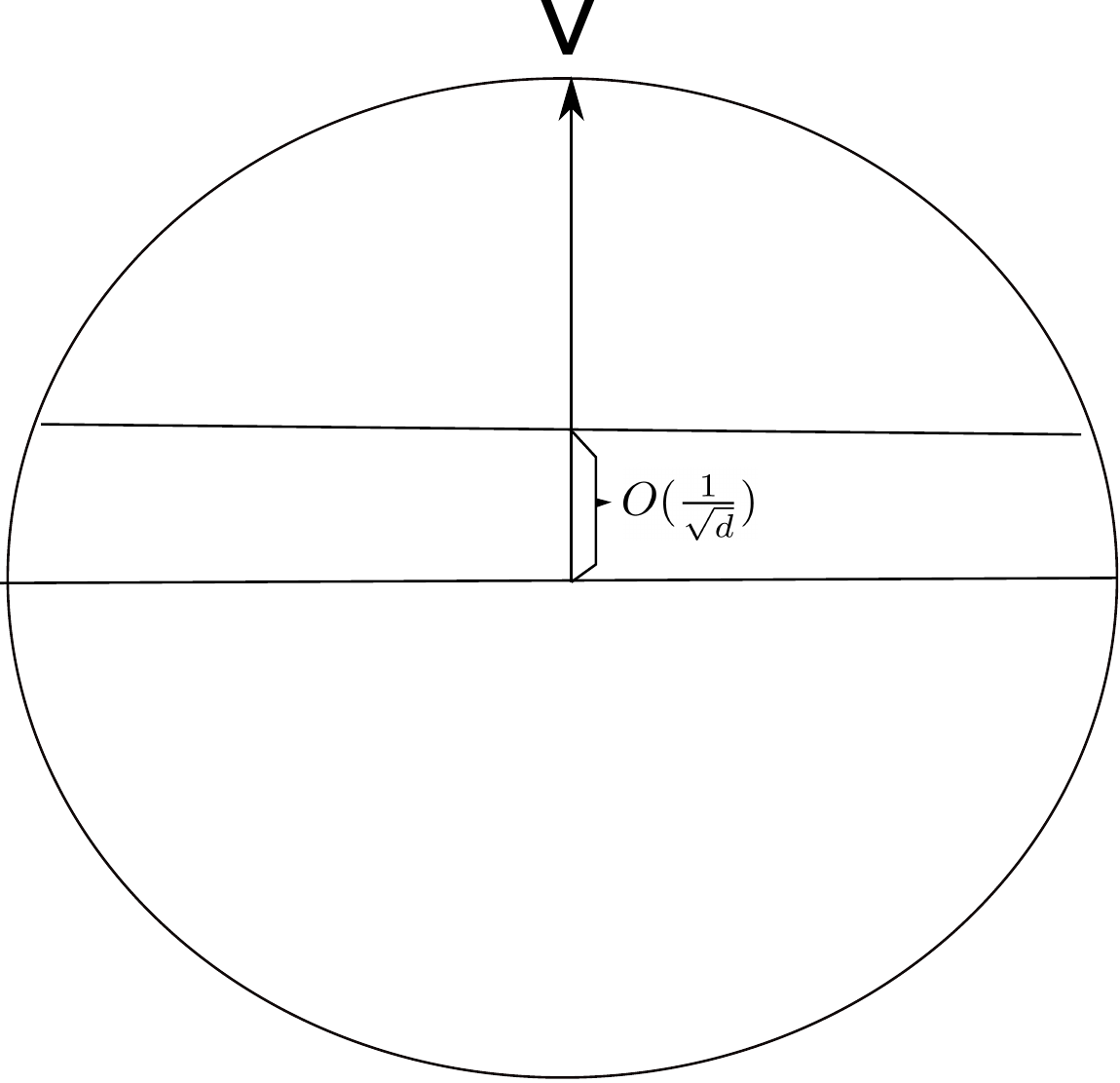}
\caption{Within a unit ball, most of the volume lies in a thin slab whose width is $O(\frac{1}{\sqrt{d}})$~\cite{blum2020foundations}.}
\label{fig:equator}
\end{figure}

Another interesting property of a high dimension space is that most of the volume of a unit ball
is concentrated near its ``equator''~\cite{blum2020foundations},
as shown in Figure~\ref{fig:equator}. More precisely, for any unit-length vector $v$ defining the ``north'', most of the volume of the unit ball
lies in the thin slab of points whose dot-product with $v$ has magnitude $O(\frac{1}{\sqrt{d}})$~\cite{blum2020foundations}.

An immediate conclusion is that for any sample from a Gaussian, it is orthogonal to most of other samples from the same Gaussian.
This is evident if we note that the dot product of any two samples tend to be $1/\sqrt{d}$, which approaches to zero with a large $d$.
Combining the Gaussian annulus theorem, we can see that samples of a high-dimensional Gaussian are mostly scattered
across direction rather than length. In other words, \textbf{direction is more important than magnitude in
a high dimensional space.} In fact, the importance of direction in high-dimensional space has been
noticed by researchers in various domains. For example, it is well-known that
the cosine distance is a better metric compared to the Euclidean distance in text analysis and information retrieval~\cite{salton1989automatic,chowdhury2010introduction,xing2015normalized}.
The same observation was also reported in speaker recognition~\cite{dehak2009support,kenny2010bayesian}.

It is worth noting that all the above conclusions are based on Gaussian distributions. If the data itself is spherical in nature,
a directional distribution will be naturally preferred, for example the
Von Mises-Fisher (VMF) distribution.
More information about directional distributions can be found in~\cite{sra2018directional,mardia2009directional}.

\section{Simulation for NL behavior}
\label{sec:sim}

In this section, we perform a simulation study to investigate the behavior of the NL score
when the data perfectly match the model assumption. We will assume a linear Gaussian model
for the NL score, and the data are sampled from the assumed model.
The first part simulates the situation where the class means have been known, and the second part
simulates the situation where the class means need to be estimated from enrollment samples.

\subsection{Known means}

In this experiment, we sample 600 classes (to represent 600 speakers) by $\pmb{\mu} \sim N(0, \mathbf{I}\pmb{\epsilon}^2 )$,
and for each class, sample 500 samples (to represent speaker vectors) by $\textbf{x} \sim N(\pmb{\mu}, \sigma^2 \mathbf{I})$ for enrollment.
By this setting, the class means can be estimated almost precisely.
For each class, we sample 30 samples as the test data, again by  $\textbf{x} \sim N(\pmb{\mu}, \sigma^2 \mathbf{I})$.
The between-class standard deviation (STD) $\pmb{\epsilon}$ is set to be 1.0 on each dimension, and the
within-class STD $\sigma$ varies from 0.1 to 5.0. The dimension varies from 10 to 80.
We evaluate the performance in terms of equal error rate (EER) for the SV task, and
identification rate (IDR) for the SI task.

\begin{figure}[htbp]
\centering\includegraphics[width=\linewidth]{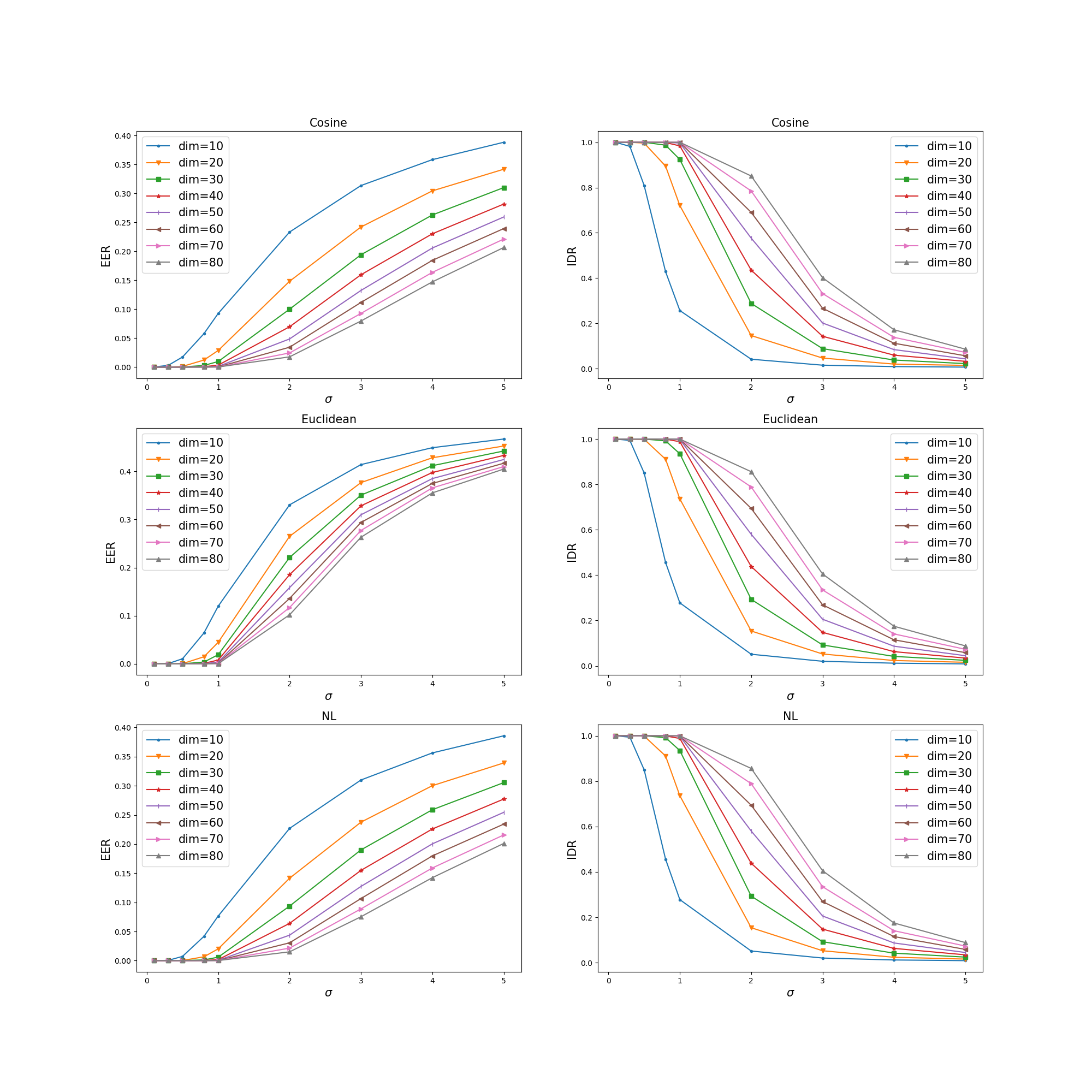}
\caption{EER (left) and IDR (right) with cosine score (top), Euclidean score (middle) and NL score (bottom), with known class means.
In each picture, different curves represent results with different dimensions. }
\label{fig:full-center}
\end{figure}

The results are shown in Figure~\ref{fig:full-center}. Firstly, it can be seen that with more dimensions, the performance in terms of both EER and IDR
is significantly improved. This is expected as a higher dimension offers better class separation as discussed in Section~\ref{sec:remark}.
In addition, with a larger within-class variance, the
performance is reduced significantly. This is also expected as a larger within-class variance leads to more overlap among classes.
Comparing the three scores, For the EER results, the cosine score can approximate the NL score with a good accuracy in terms of EER,
though the Euclidean score performs relatively worse.
For the IDR results, the three scores are
largely comparable. In particular, the Euclidean score obtains exact the same performance as the NL score.

\begin{figure}[htbp]
\centering\includegraphics[width=\linewidth]{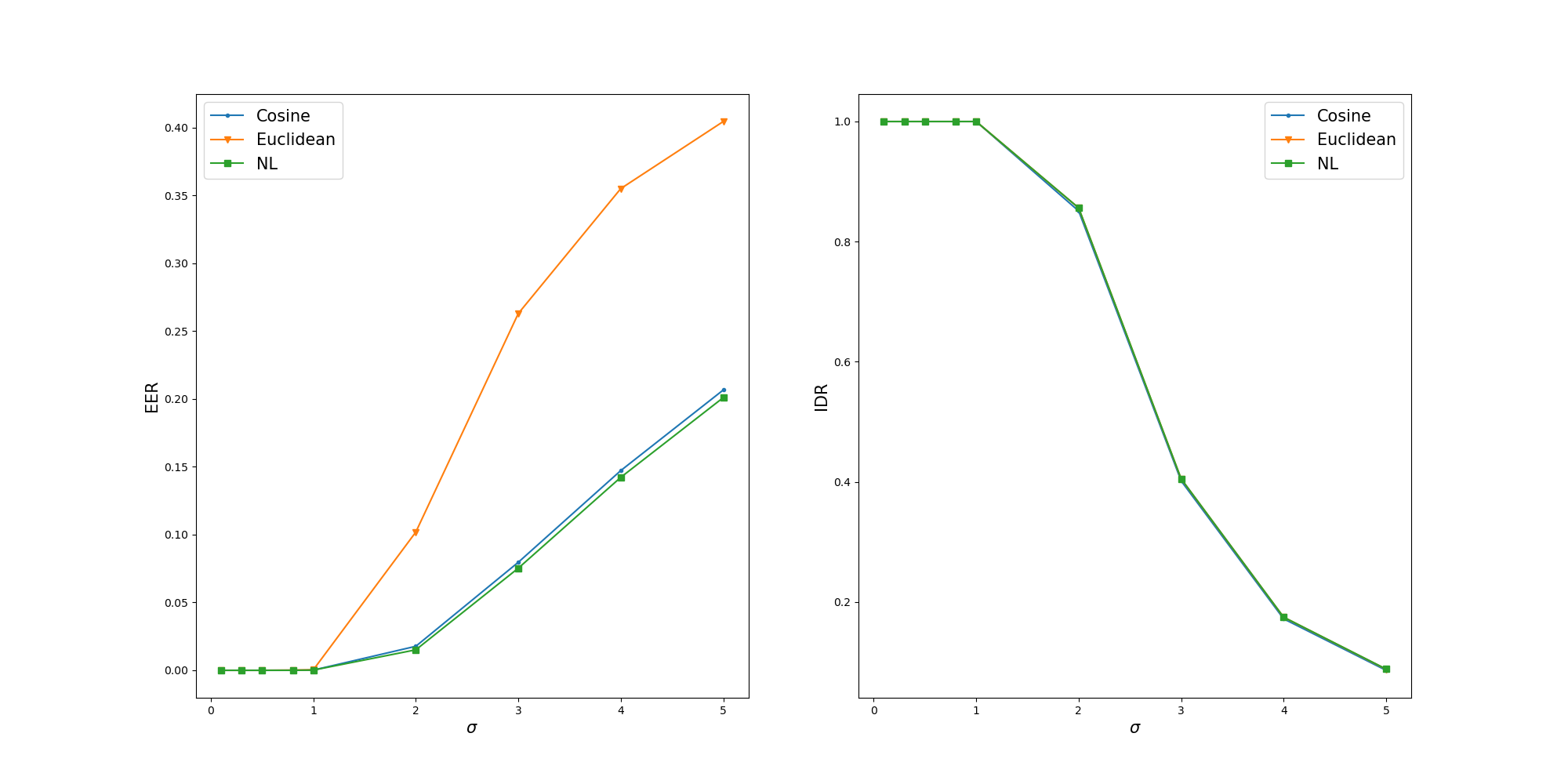}
\caption{EER (left) and IDR (right) with cosine score, Euclidean score and NL score, within known means.
In each picture, different curves represent results with different scores. The dimension is set to be 80. }
\label{fig:full-center-dim80}
\end{figure}

For a better comparison on the three scores, we draw their performance in the same picture, where we choose the strongest system with 80 dimensions.
The results are shown in Figure~\ref{fig:full-center-dim80}. It demonstrates that for the SI task (IDR results), the three scores perform nearly the same. For the SV task (EER results),
The Euclidean score performs very bad, especially when the within-covariance is large.
The NL score is slightly better than the cosine score on SV, though the difference is marginal.

\begin{figure}[htbp]
\centering\includegraphics[width=\linewidth]{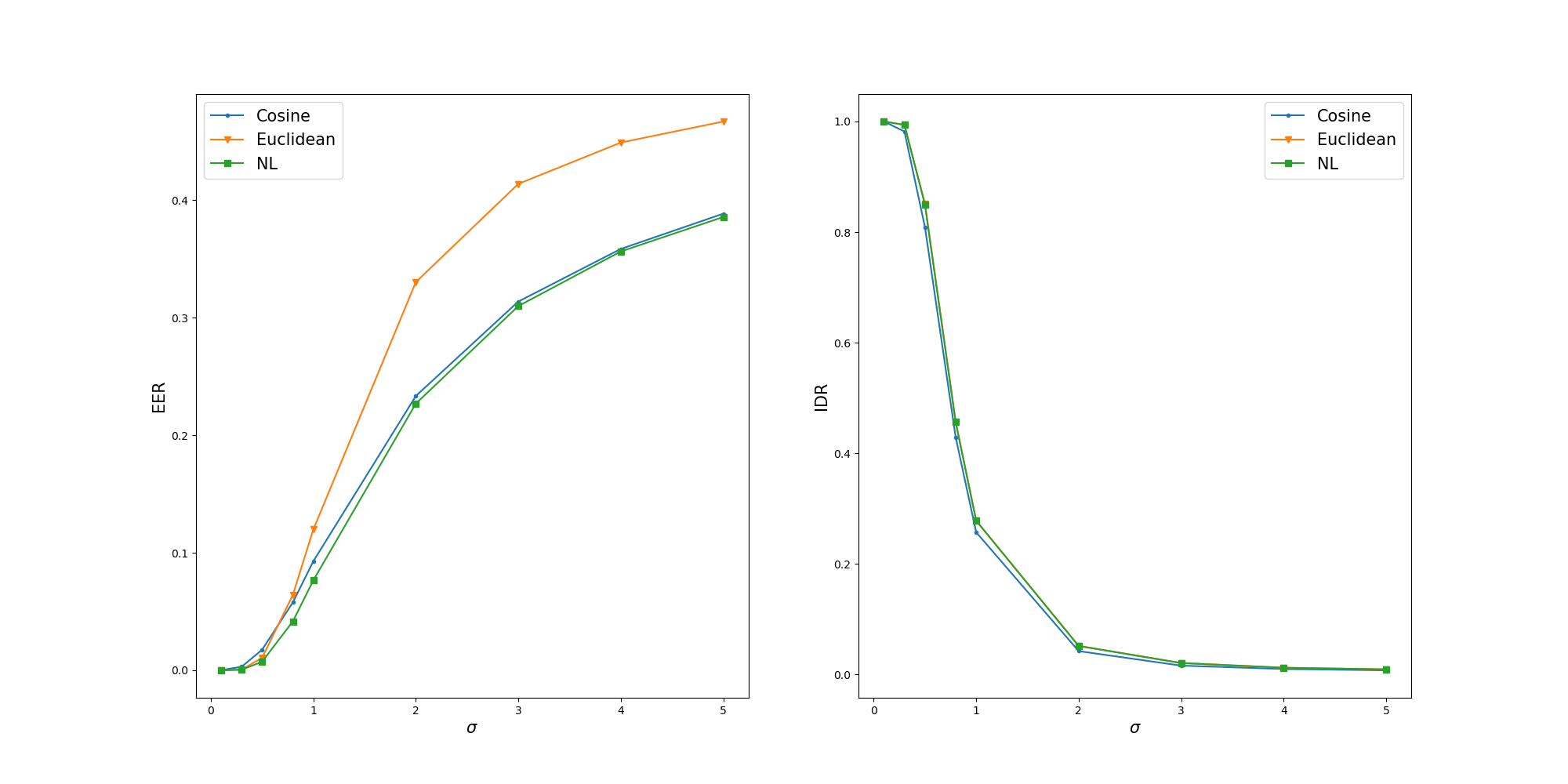}
\caption{EER (left) and IDR (right) with cosine score, Euclidean score and NL score, with known means.
In each picture, different curves represent results with different scores. The dimension is set to be 10.}
\label{fig:full-center-dim10}
\end{figure}

Similarly, we plot the results with the dimension set to be 10, as shown in Figure~\ref{fig:full-center-dim10}. It can be seen that in this more
tough situation, NL shows some advantage on the SI task compared to the cosine score. On the SV task,
the cosine score matches the NL score better when the within-class variance is large, though the Euclidean score matches the NL score when the
within-class variance is very small. All these results are expected, and in accordance with the prediction by Eq.(\ref{eq:nl-known}).

\subsection{Unknown means}

\begin{figure}[htbp]
\centering\includegraphics[width=\linewidth]{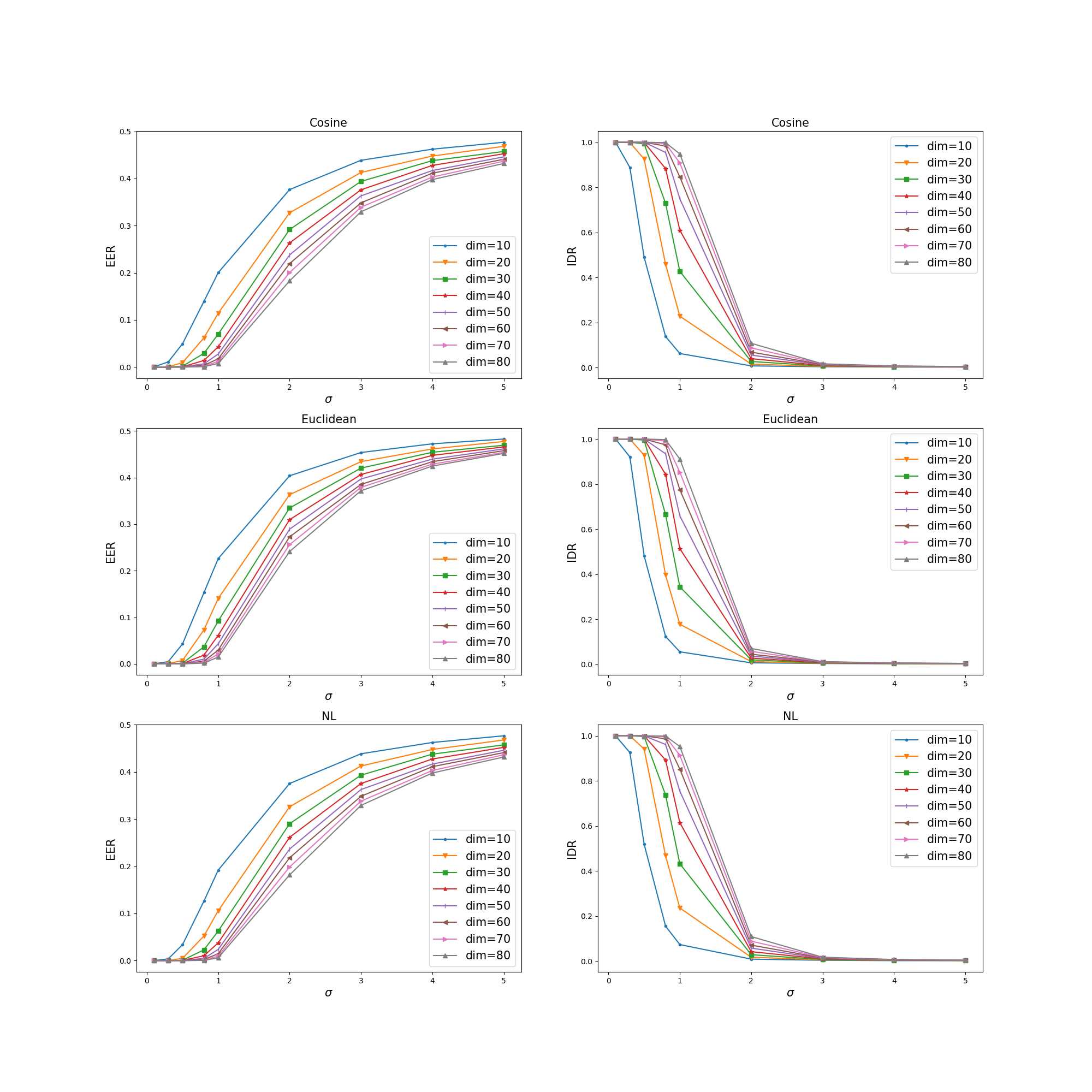}
\caption{EER (left) and IDR (right) with cosine score (top), Euclidean score (middle) and NL score (bottom), with unknown class means.
In each picture, different curves represent results with different dimensions. }
\label{fig:part-center}
\end{figure}

In the second experiment, we study the behavior of the NL score with unknown class means. Again, we sample 600 classes; for each class, we sample 1 sample for enrollment and 3 samples
for test. Each test runs 500 rounds, and the average performance is reported to ensure the statistical significance.
Other settings are the same as in the known-mean experiment. The results are shown in Figure~\ref{fig:part-center}. It can be seen that with unknown means, the
overall performance with any score is worse than that reported in the known-mean experiment (Figure~\ref{fig:full-center}). This impact is much more obvious
in terms of IDR: with a slightly larger within-class variance, the performance is reduced substantially. Note that in this case, the Euclidean score does not match the
NL on the SI task, due to the compensation factor as shown in Eq.~\ref{eq:posterior-unknown}.

\begin{figure}[htbp]
\centering\includegraphics[width=\linewidth]{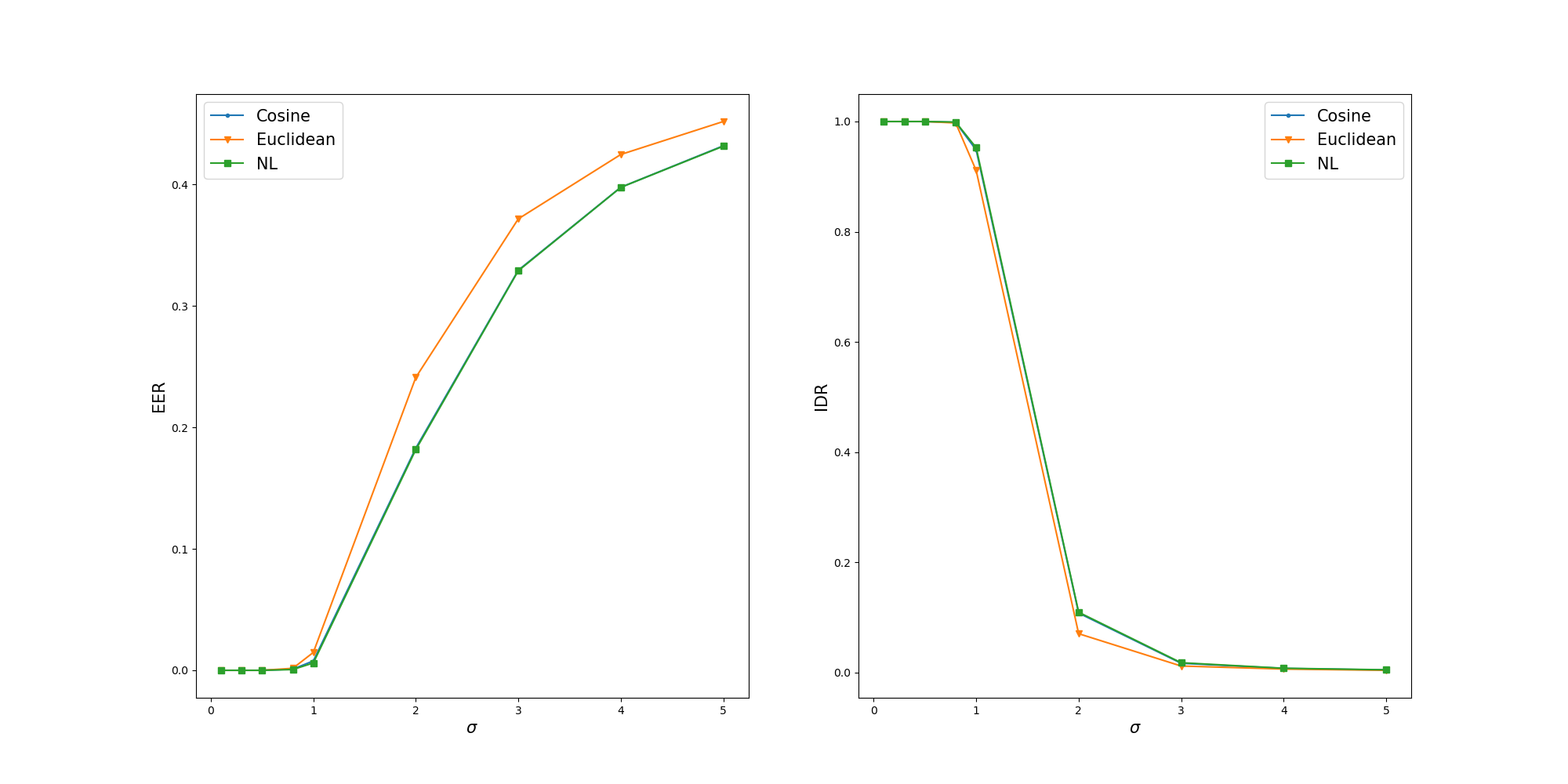}
\caption{EER (left) and IDR (right) with cosine score, Euclidean score and NL score, with unknown means.
In each picture, different curves represent results with different scores.  The dimension is set to be 80.}
\label{fig:part-center-dim80}
\end{figure}

\begin{figure}[htbp]
\centering\includegraphics[width=\linewidth]{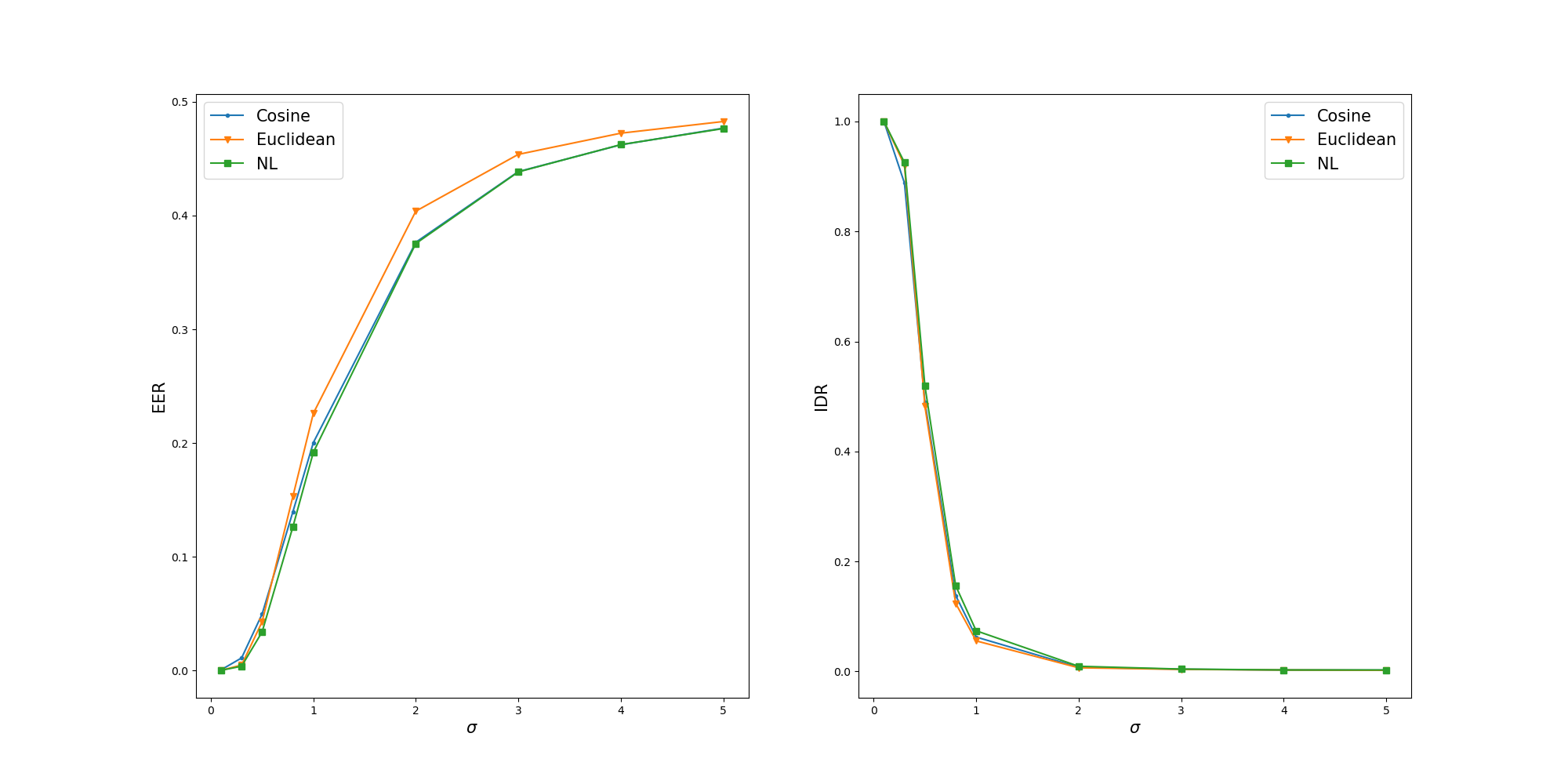}
\caption{EER (left) and IDR (right) with cosine score, Euclidean score and NL score, with unknown means.
In each picture, different curves represent results with different scores.  The dimension is set to be 10.}
\label{fig:part-center-dim10}
\end{figure}

Again, we plot the results with the three scores in the same picture, as shown in Figure~\ref{fig:part-center-dim80} and Figure~\ref{fig:part-center-dim10},
where the dimension is set to 80 and 10, respectively.
Firstly we find that the cosine score offers a rather good approximation in most cases, for both the SV and SI tasks, while the Euclidean score does not.
Moreover, the cosine score matches the NL score better on the SV task when the within-class variance is large. This conclusion is the same as the one obtained in the known-mean experiment,
and is consistent with the quantitative analysis based on Eq.(\ref{eq:nl-unknown}).

\section{Discussion}

An important merit that the NL scoring approach offers is an analytical perspective for the scoring process.
According to the NL formula, the optimal score is determined by three components:
the enrollment model $p(\pmb{\mu}|\pmb{x}_1, ..., \pmb{x}_{n_k})$,
the prediction model $p(\pmb{x}|\pmb{\mu})$, and the normalization
model $p(\pmb{x})$. If all these three components describe the data correctly, the score will be optimal.
Vice versa, mismatch between the data and any of the three components will cause suboptimum.
The mismatch could be:
(1) mismatch on distribution type (e.g., Gaussian assumed but Laplacian in reality);
(2) mismatch on the mean (mean shift); (3) mismatch on the covariance (covariance drift).
This analytical view provides a powerful tool by which we can analyze how the performance reduction is caused by
a particular imperfection, and design suitable algorithms to compensate for the impact.
Recently, using this tool we provided a theoretically optimal solution for the enroll-test mismatch problem and 
achieved a big success~\cite{li2020match,li2020squeeze}.

Although our discussion focuses on the linear Gaussian NL,
the NL formulation is general and can be easily extended by using nonlinear and non-Gaussian models,
so that it deal with more complex data.
Recently, we provide such an extension~\cite{li2020nda}, by applying the invariance property of
the NL score under invertible transforms, as discussed in Section~\ref{sec:remark}.
Specifically, we learn an invertible transform that maps the original data to
a latent space where the data can be modeled by a linear Gaussian.
According to the equivalence of the NL score in the original and the transformed space,
this transform allows us using a linear Gaussian NL model to score data with a complex distribution.
This is essentially a nonlinear extension of the PLDA model, which we call \emph{neural discriminant
analysis (NDA)}. In our previous study, the NDA model produces very promising results~\cite{li2020nda}.

The MBR optimum with NL scoring encourages more research on the speaker embedding approach.
Since we have known that the NL score is MBR optimal, its performance will be ensured if the distribution of the speaker
vectors meet the assumption of the model.
This performance ensurance represents a clear advantage of the embedding approach compared to the so-called end-to-end
approach~\cite{heigold2016end,zhang2017end,Chowdhury17attention}.
Moreover, since the NL score is optimal if and only if the speaker vectors follow the assumed generative model,
more research is encouraged on normalizing the speaker vectors, rather than
pursuing other complicated scoring methods (e.g., discriminative PLDA~\cite{burget2011discriminatively})
or score calibration~\cite{leeuwen2013the,Cumani2019tied}.
Our recent work shows that speaker vector normalization is highly promising, especially when a flow-based deep 
generative neural net is used~\cite{cai2020deep}. We name this model as \emph{discriminative normalization flow} (DNF).
It was shown that DNF can be regarded as a deep generative LDA~\cite{cai2020lda}, and can effectively 
normalize class distributions to Gaussians. Moreover, by employing a maximum Gaussian training,
the normalization is even more successful~\cite{cai2020mg}.

\section{Conclusions}
\label{sec:con}

We present an analysis on the optimal score for speaker recognition based on the MAP principle.
The analysis shows that the normalized likelihood (NL) is optimal for both identification and
verification tasks in the sense of minimum Bayes risk. We also show that if the underlying generative
model is linear Gaussian, the NL score is equivalent
to the popular PLDA LR, and the cosine score and Euclidean score can be regarded as two approximations of the
optimal NL score. Some properties of the NL score were discussed, and a simple simulation experiment was
performed to understand the behavior of the NL score. More complicated simulations that take into account 
the configurations of real speaker recognition systems can be found in our recent publication~\cite{wang2020simulation}.

\section*{Acknowledgement}

Thanks to Dr. Lantian Li, Yunqi Cai and Zhiyuan Tang for the valuable discussion.

\bibliographystyle{plainnat}
\bibliography{refs}

\end{document}